\documentclass[journal]{IEEEtran}
\usepackage{amsmath,amsfonts}
\usepackage{algorithmic}
\usepackage{algorithm}
\usepackage{array}
\usepackage{textcomp}
\usepackage{url}
\usepackage{verbatim}
\usepackage{graphicx}
\usepackage[numbers]{natbib}
\usepackage{subcaption} 
\usepackage{booktabs, tabularx}
\usepackage{capt-of}
\usepackage[colorlinks=true, linkcolor=blue]{hyperref}
\usepackage{makecell} 
\begin{document}

\title{VRUD: A Drone Dataset for Complex Vehicle-VRU Interactions within Mixed Traffic}


\author{Ziyu Wang\thanks{This work was supported in part by the Science and Technology Development Program of Jilin Province under Grant 20240302052GX and in part by the National Natural Science Foundation of China under Grant 52075213.}%
\thanks{Ziyu Wang is with the National Key Laboratory of Automotive Chassis Integration and Bionics, Jilin University, Changchun 130025, China, and also with DRIVEResearch, Changchun, China (e-mail: ziyu25@mails.jlu.edu.cn).}, 
Hongrui Kou\thanks{Hongrui Kou is with the National Key Laboratory of Automotive Chassis Integration and Bionics, Jilin University, Changchun 130025, China (e-mail: kouhr23@mails.jlu.edu.cn).}, 
Cheng Wang\thanks{Cheng Wang is with the School of Engineering and Physical Sciences, Heriot-Watt University, Edinburgh EH1 4AS, U.K. (e-mail: Cheng.Wang@hw.ac.uk).}, 
Ruochen Li, 
Hubert P. H. Shum, 
Amir Atapour-Abarghouei\thanks{Ruochen Li, Hubert P. H. Shum, and Amir Atapour-Abarghouei are with the Department of Computer Science, Durham University, Durham DH1 3LE, U.K. (e-mail: ruochen.li@durham.ac.uk; hubert.shum@durham.ac.uk; amir.atapour-abarghouei@durham.ac.uk).}, 
Yuxin Zhang$^{*}$\thanks{$^{*}$Yuxin Zhang is the corresponding author. He is with the National Key Laboratory of Automotive Chassis Integration and Bionics, Jilin University, Changchun 130025, China; the Department of Computer Science, Durham University, Durham DH1 3LE, U.K.; and DRIVEResearch, Changchun, China (e-mail: yuxinzhang@jlu.edu.cn).}}

\IEEEaftertitletext{%
\begin{center}
\includegraphics[width=\textwidth]{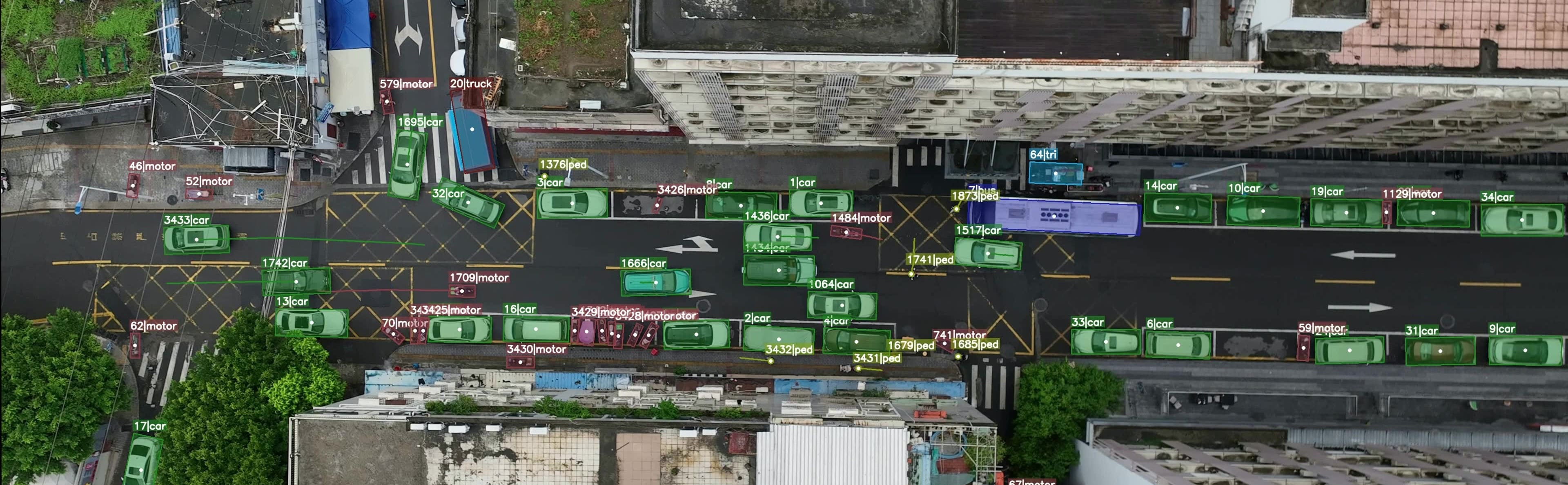}
\captionof{figure}{Example of a recorded sequence showing bounding boxes and labels for detected traffic participants. The bounding box color indicates the category of each traffic participant, with historical trajectories rendered in the corresponding color. Each traffic participant is assigned a unique identifier).}
\label{fig:trackresults}
\end{center}
}

\maketitle

\begin{abstract}
The Operational Design Domain (ODD) of urban-oriented Level 4 (L4) autonomous driving, especially for autonomous robotaxis, confronts formidable challenges in complex urban mixed traffic environments. These challenges stem mainly from the high density of Vulnerable Road Users (VRUs) and their highly uncertain and unpredictable interaction behaviors. However, existing open-source datasets predominantly focus on structured scenarios such as highways or regulated intersections, leaving a critical gap in data representing chaotic, unstructured urban environments. To address this, this paper proposes an efficient, high-precision method for constructing drone-based datasets and establishes the Vehicle-Vulnerable Road User Interaction Dataset (VRUD), as illustrated in Figure \ref{fig:trackresults}. Distinct from prior works, VRUD is collected from typical ``Urban Villages'' in Shenzhen, characterized by loose traffic supervision and extreme occlusion. The dataset comprises 4 hours of 4K/30Hz recording, containing 11,479 VRU trajectories and 1,939 vehicle trajectories. A key characteristic of VRUD is its composition: VRUs account for about 87\% of all traffic participants, significantly exceeding the proportions in existing benchmarks. Furthermore, unlike datasets that only provide raw trajectories, we extracted 4,002 multi-agent interaction scenarios based on a novel Vector Time to Collision (VTTC) threshold, supported by standard OpenDRIVE HD maps. This study provides valuable, rare edge-case resources for enhancing the safety performance of ADS in complex, unstructured urban environments.

To facilitate further research, we have made the VRUD dataset open-source at: \url{https://zzi4.github.io/VRUD/}

\end{abstract}


\section{Introduction}

The iterative evolution of Autonomous Driving Systems (ADS) is fundamentally anchored in the availability of massive, high-fidelity real-world traffic data \cite{li2020trajectory}. Throughout the ADS lifecycle, such data is pivotal. During the development phase, it serves as a critical foundation for developing trajectory prediction \cite{li2025vite, li2025bp} and path planning modules, ensuring system logic aligns with human decision-making patterns \cite{10736349,9826385}. In the testing phase, it functions as ground truth data for algorithm validation and supports the construction of diverse test cases, which are essential for delineating the performance boundaries of the system \cite{10321740}. For the emerging end-to-end architectures, the volume and authenticity of data directly determine the upper limit of system performance \cite{chen2024end, le2022survey}. However, the most critical challenge currently lies in urban mixed traffic environments. Unlike structured scenarios (e.g., highways) with clear rules and broad visibility, mixed traffic is characterized by significant complexity and uncertainty \cite{lin2025safety, garg2023can}. These environments are characterized by high VRU density, frequent occlusions, and irregular traffic behaviors such as jaywalking and wrong-way driving \cite{muduli2023prediction}. The stochastic nature of VRU behaviors and the lack of signal communication jointly make accurate trajectory prediction and safe interaction particularly challenging \cite{kutela2022mining}. Consequently, capturing real-world interaction behaviors in these chaotic environments is not merely a data collection task but a critical prerequisite for deploying high-level ADS in complex urban roads \cite{harkin2024vulnerable,yu2021automated}.

Despite the critical necessity of such data, the current landscape of open-source datasets reveals a significant scarcity regarding complex, unstructured urban environments. While representative datasets such as KITTI \cite{geiger2013vision} and nuScenes \cite{caesar2020nuscenes} have propelled the field forward, they predominantly focus on structured roads dominated by vehicle-vehicle interactions. Although recent drone-based datasets like highD \cite{krajewski2018highd} and INTERACTION \cite{zhan2019interaction} offer bird's-eye views, they largely target highways or regulated intersections with relatively ordered traffic flows. Thus, there is a distinct lack of datasets that capture the ``long-tail'' scenarios found in densely populated, unregulated areas, such as ``Urban Villages'', where VRUs constitute the majority of traffic participants and interactions are driven by negotiation rather than rigid rules. This data gap restricts the ability of current ADS to develop socially compliant driving strategies for safe navigation in shared environments.

To bridge this gap, this paper introduces VRUD, a drone-based dataset designed to capture complex vehicle-VRU interactions within mixed traffic. Collected from typical urban villages in Shenzhen, VRUD serves as a foundational resource for analyzing unstructured behavioral habits, training interaction-aware prediction models, and validating the safety and robustness of end-to-end ADS in highly dynamic environments.

The main contributions of this work are summarized as follows:
\begin{itemize}
    \item \textbf{High-Density VRU Dataset:} We construct a large-scale, high-resolution dataset characterized by extreme VRU density (comprising nearly 87\% of participants) and diverse mixed traffic behaviors.
    \item \textbf{Standardized Scenario Library:} Beyond providing raw high-precision trajectories, we propose a comprehensive data processing pipeline that extracts multi-agent interaction scenarios. This is supported by OpenDRIVE HD maps and a novel Vector Time-to-Collision (VTTC) threshold, which extends conventional TTC by incorporating vector velocity and relative position to quantify interaction intensity.
    \item \textbf{Behavioral Extraction \& Analysis:} We provide a detailed statistical analysis of unique VRU interaction patterns, offering test cases to support and evaluate downstream ADS tasks.
\end{itemize}

\begin{figure*}[htbp]  
    \centering  
    \begin{subfigure}[b]{0.48\textwidth}
        \centering
        \includegraphics[width=\textwidth]{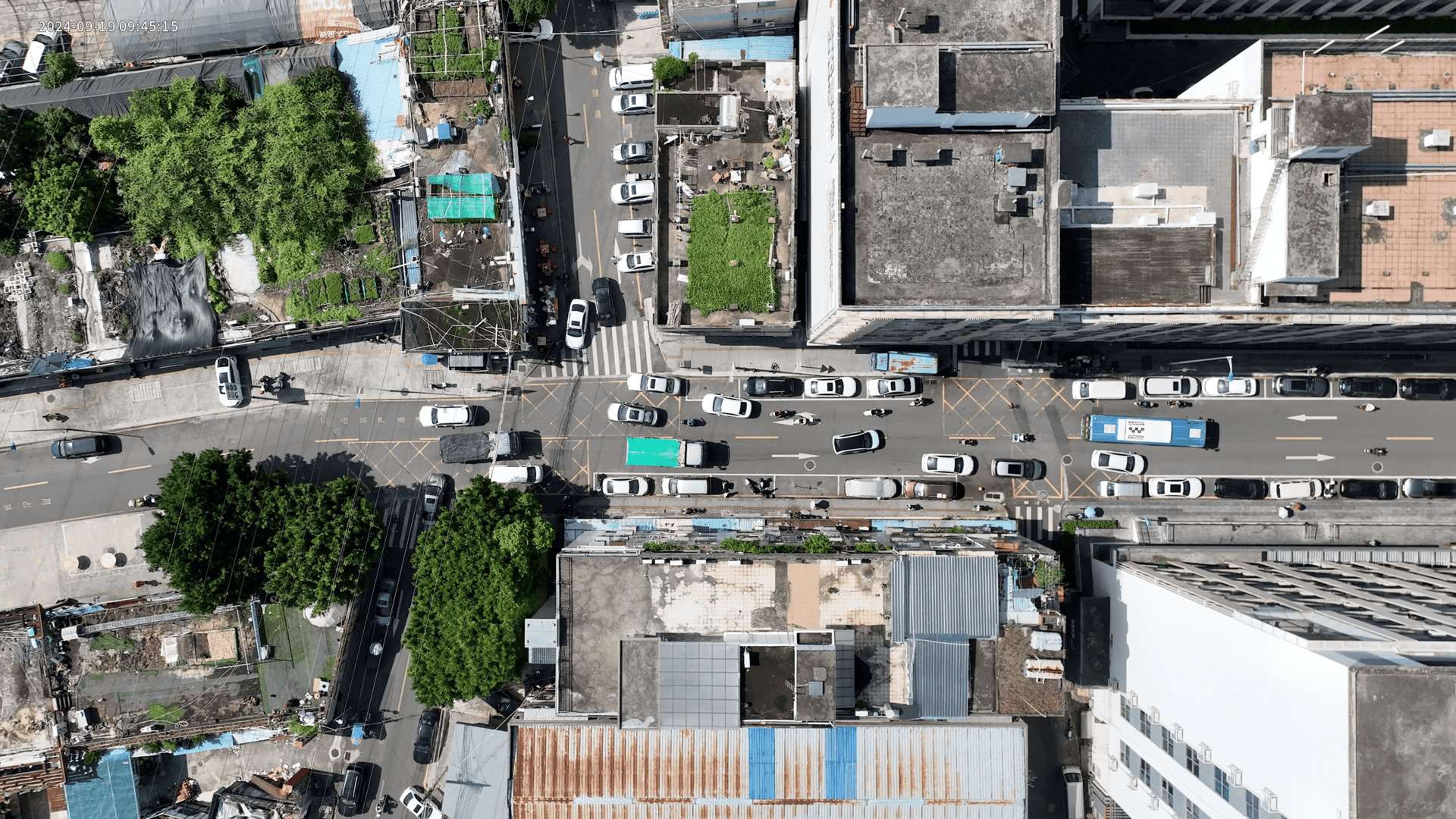} 
        \caption{}
        \label{fig:suba4}  
    \end{subfigure}
    \hfill 
    \begin{subfigure}[b]{0.48\textwidth}
        \centering
        \includegraphics[width=\textwidth]{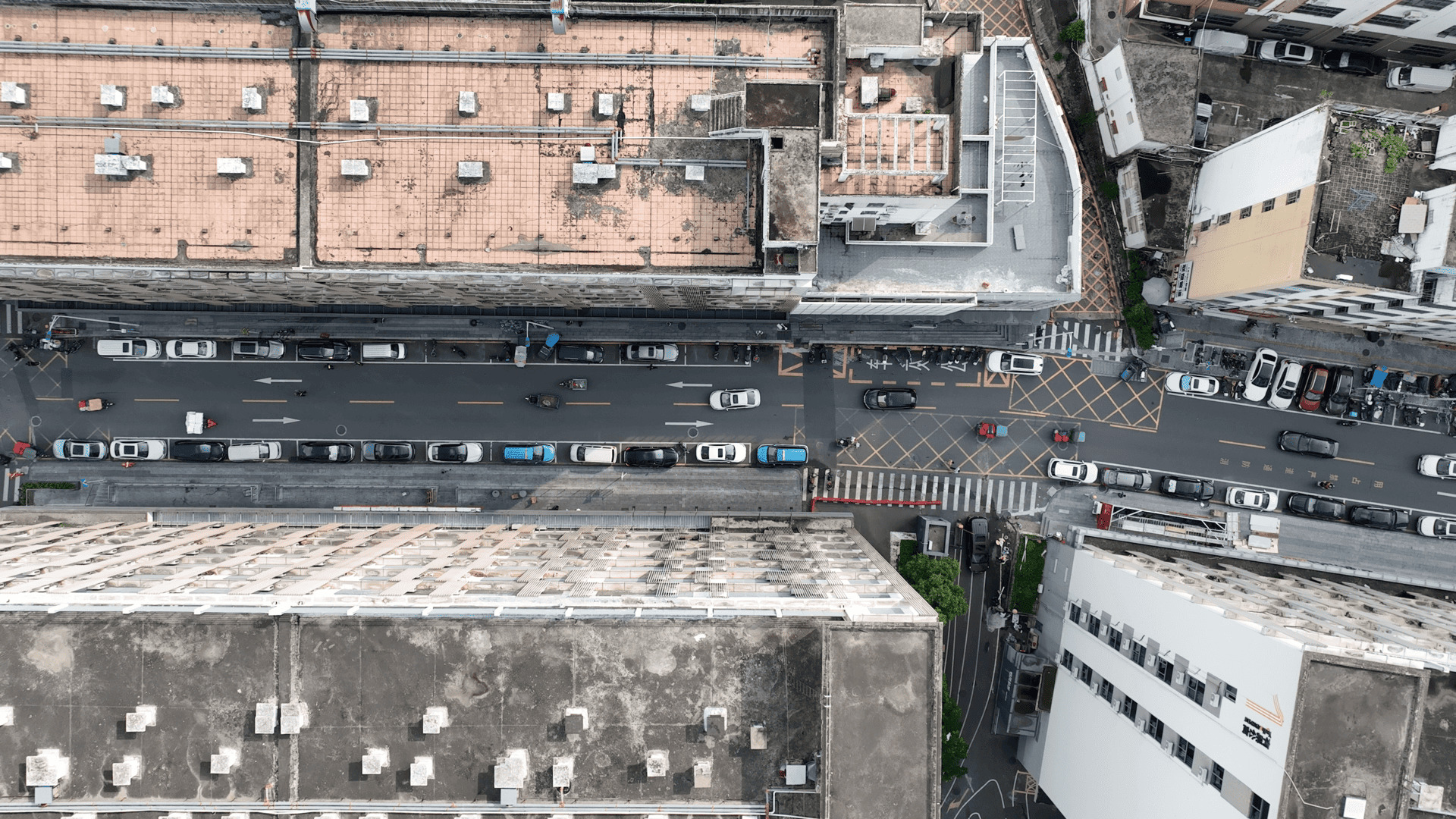} 
        \caption{}
        \label{fig:subb4}
    \end{subfigure}
    \caption{Two Data Collection Sites for VRUD. (a) depicts an irregular intersection, both featuring two-way single-lane traffic. The four areas divided at the intersection correspond to two snack streets and two residential apartment complexes. (b) shows a two-way single-lane road, with residential apartments lining both sides. At both sites, apartment entrances and bus stops are situated along the road edges. In addition, the parking spaces on either side of the roads are occupied by long-term parked vehicles.}
    \label{fig:sites}  
\end{figure*}

\section{Related Work}

Building upon our previous comprehensive survey \cite{wang2025comprehensive}, we systematically categorized dataset construction in autonomous driving into three mainstream paradigms: vehicle-based, roadside-based, and drone-based collection. Vehicle-based datasets (e.g., nuScenes \cite{caesar2020nuscenes}, KITTI \cite{geiger2013vision}) and roadside datasets (e.g., NGSIM \cite{NGSIM2016}, I-24 MOTION \cite{gloudemans202324}) have laid the foundation for data-driven research. However, a comparative analysis of technical attributes reveals inherent limitations in these methods regarding field-of-view (FOV) coverage and occlusion handling. In contrast, drone-based collection (e.g., highD \cite{krajewski2018highd}, AD4CHE \cite{zhang2023ad4che}) demonstrates superior adaptability for low-altitude mixed traffic scenarios. Particularly in environments with high traffic density, the aerial perspective offers distinct advantages, including occlusion-free data acquisition, non-interference with natural traffic flow, and high operational efficiency, making it the optimal choice for capturing complex interactions.

Leveraging aerial photography, several studies have established datasets for mixed traffic scenarios. Notable examples include the rounD \cite{krajewski2020round} dataset for roundabouts, and intersection datasets such as inD \cite{bock2020ind}, INTERACTION \cite{zhan2019interaction}, and SIND \cite{xu2022drone}. While these datasets incorporate diverse participants including VRUs, their collection sites are predominantly located in structured road environments with relatively regulated traffic flows. They generally lack the chaotic characteristics of unregulated residential zones—scenarios defined by lower supervision intensity, irregular intersections, and complex elements like roadside stalls. Consequently, open-source datasets targeting these unstructured, high-density mixed traffic environments remain scarce, limiting the validation of ADS in ``long-tail'' urban scenarios.

Existing datasets focusing on pedestrian-vehicle interactions remain limited in both spatial resolution and participant diversity. For instance, the TUMDOT–MUC \cite{kutsch2024tumdot} dataset covers mixed road sections but suffers from excessive flight altitude, resulting in insufficient resolution to distinguish individual pedestrians within crowds or capture fine-grained behavioral cues. Similarly, datasets like CITR\&DUT \cite{yang2019top} and Campus \cite{robicquet2016learning} focus on pedestrian-vehicle interactions but are restricted to campus environments. These scenarios exhibit a homogeneity of participants (mostly students) whose behavioral habits do not reflect the heterogeneity of the general population. Furthermore, they lack diverse non-motorized VRUs, such as delivery tricycles and motorcycles, which are prevalent in real-world urban traffic.


The remainder of this paper is organized as follows: Section \ref{sec:construction} details the methodology for dataset construction, including parameter configuration, preprocessing, and validation. Section \ref{sec:statistics} presents the statistical analysis of the dataset and characterizes the extracted interaction behaviors. Finally, Section \ref{sec:conclusion} concludes the paper and outlines future work.

\section{Construction of dataset}
\label{sec:construction}
This section is organized into several subsections, covering data acquisition parameter configuration, preprocessing, trajectory acquisition, postprocessing, and behavior extraction.
\subsection{Data acquisition parameter configuration}

To acquire the interaction behaviors between VRUs and vehicles more efficiently and to adapt to the complex urban mixed traffic environment, we selected two irregular intersections near residential areas in an urban village in Shenzhen, China. The traffic participants in this area are highly diverse, including buses, ride-hailing vehicles, food delivery electric bikes, and pedestrians. The area includes bus stops, residential buildings, and snack streets; yet, it lacks traffic surveillance cameras. The actual scene is illustrated in Figure \ref{fig:sites}. To ensure the accurate acquisition of behavioral data for small targets such as pedestrians and non-motorized vehicles, we set the camera resolution to 4K, the drone flight altitude to 80 meters, and the video frame rate to 30 fps. Data acquisition was conducted exclusively during morning and evening peak hours, resulting in a total effective duration of 4 hours.

Furthermore, all drone operations were conducted strictly within legally permitted low-altitude airspace in Shenzhen, adhering to local flight regulations regarding permissible timeframes and zones. To safeguard personal privacy, the drone was operated at a cruising altitude of 80 meters. At this specified altitude and resolution, no Personally Identifiable Information (PII), such as human faces or vehicle license plates, can be visually recognized. Consequently, the dataset is inherently anonymized, ensuring that the privacy of all traffic participants is fully protected without the need for additional facial or license plate blurring techniques.
\begin{figure}[htbp]  
    \centering  
    \begin{subfigure}[b]{\linewidth}
        \centering
        \includegraphics[width=\linewidth]{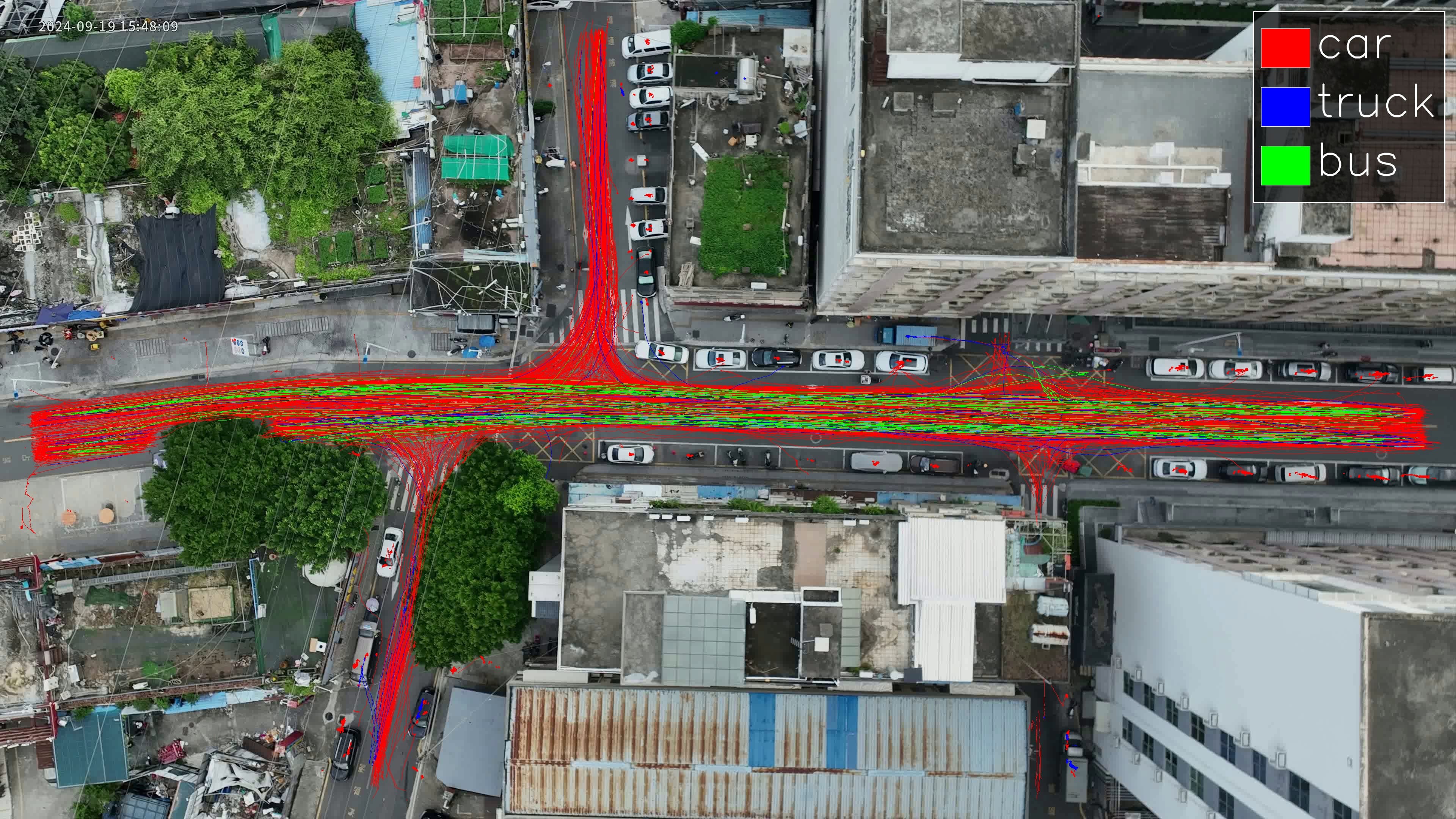} 
        \caption{}
        \label{fig:suba5}  
    \end{subfigure}
    \hfill 
    \begin{subfigure}[b]{\linewidth}
        \centering
        \includegraphics[width=\linewidth]{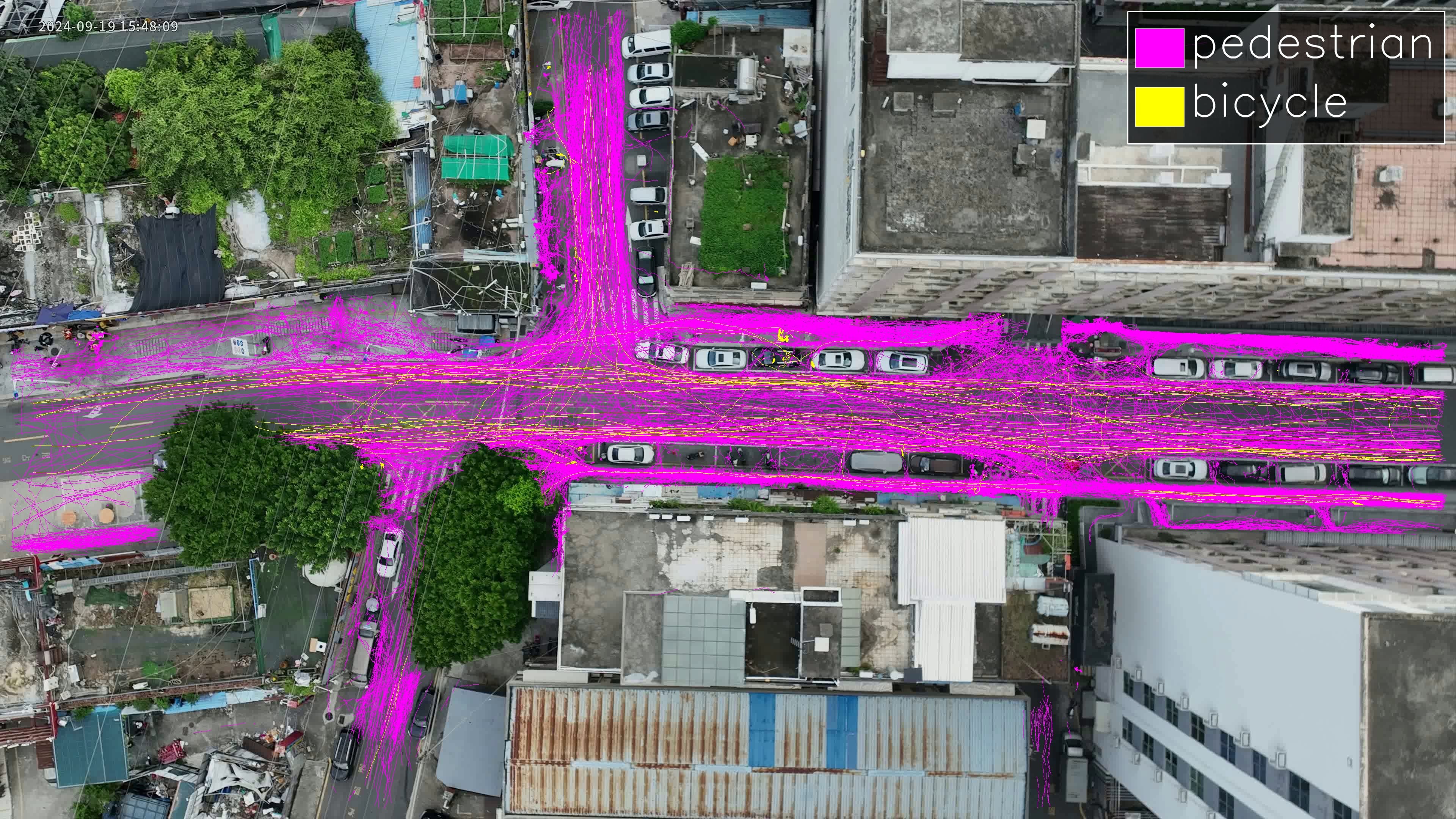} 
        \caption{}
        \label{fig:subb5}
        
    \end{subfigure}
    \hfill 
    \begin{subfigure}[b]{\linewidth}
        \centering
        \includegraphics[width=\linewidth]{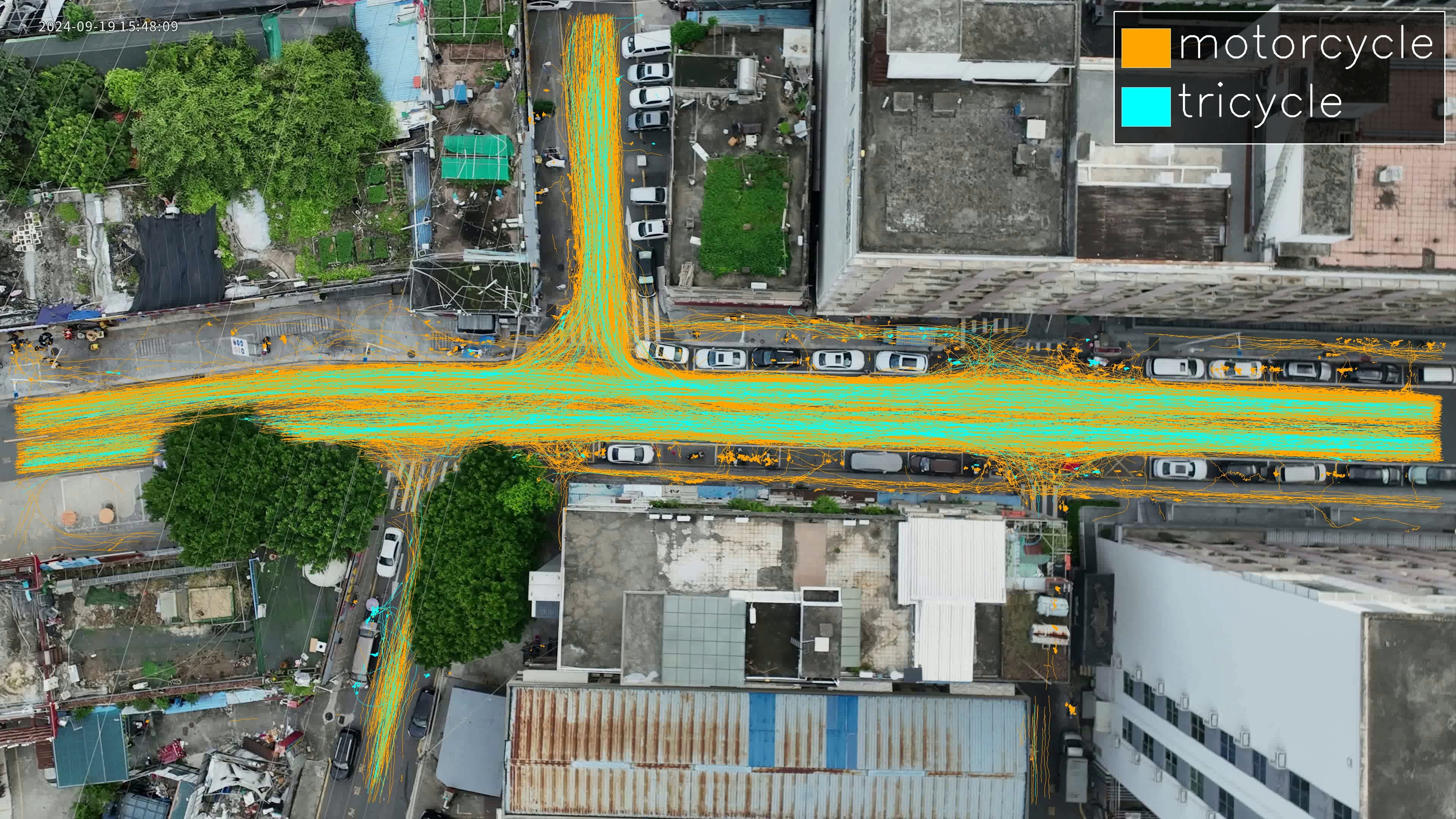} 
        \caption{}
        \label{fig:subc5}
    \end{subfigure}
    \caption{Each subplot visualizes all annotated trajectories appearing in the scene: (a) cars, buses, and trucks; (b) pedestrians and cyclists; and (c) motorcycles and tricycles. The comparison reveals that the trajectory patterns of VRUs are significantly more disordered and scattered.}
    \label{fig:trajectories}  
\end{figure}
\subsection{Data preprocessing}
To ensure the consistency of subsequent coordinate systems and the accuracy of trajectory extraction, we conducted single-video stabilization and multi-video alignment. These two steps were designed to separately address video jitter caused by drone vibrations and environmental factors, as well as field-of-view deviations resulting from multiple take-off and landing operations. 
\subsubsection{Single-video stabilization}
\begin{figure}[htbp]  
    \centering  
    \begin{subfigure}[b]{0.48\linewidth}
        \centering
        \includegraphics[width=\linewidth]{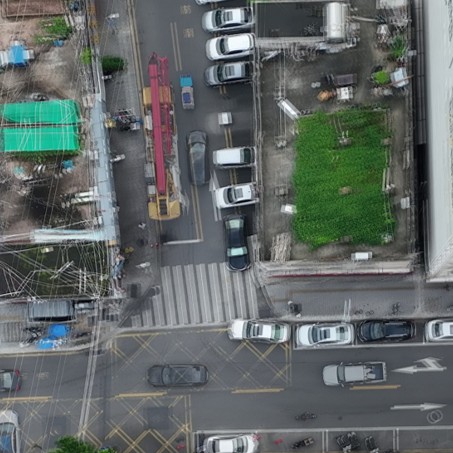} 
        \caption{}
        \label{fig:suba1}  
    \end{subfigure}
    \hfill 
    \begin{subfigure}[b]{0.48\linewidth}
        \centering
        \includegraphics[width=\linewidth]{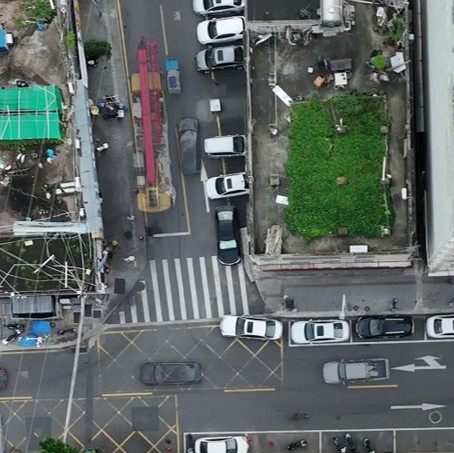} 
        \caption{}
        \label{fig:subb1}
    \end{subfigure}
    \caption{(a) shows the result of overlaying the first and last frames of a video captured in a single acquisition. (b) presents the overlay effect of the first and last frames after single-video stabilization processing. It can be clearly observed that the deviation of the background positions between the two frames has been effectively corrected, which ensures the consistency between the image coordinate system of all targets in a single subsequent video and the actual global coordinate system.}
    \label{fig:stable}  
\end{figure}

Video jitter exerts a significant negative impact on subsequent analyses. Since trajectory extraction is based on the results of image-based object detection, where all detection outputs are defined in the image coordinate system, the presence of video jitter will lead to positional deviations of stationary targets across different frames. Therefore, the core objective of single-video stabilization is to unify and rectify the coordinate systems of detection results across all frames of the video. The technical pipeline of single-video stabilization adopted in this study is as follows: Harris corner detection \cite{derpanis2004harris} is first employed to identify feature points in the video frames, and then dense optical flow \cite{lucas1981iterative} is used to calculate the transformation matrix of each frame relative to the reference frame based on the motion vectors of these feature points. The comparison before and after video stabilization is shown in Figure \ref{fig:stable}.

\subsubsection{Multi-video alignment}
\begin{figure}[htbp]  
    \centering
    \includegraphics[width=\linewidth]{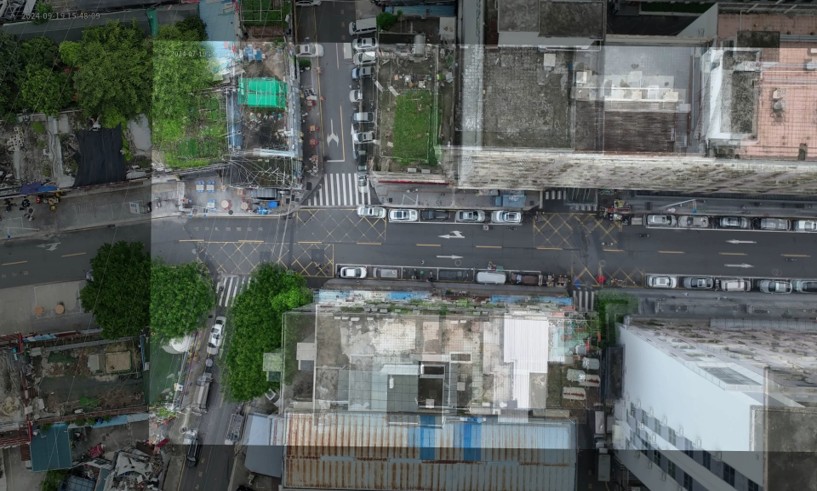}
    \caption{This figure illustrates the overlay effect of two images after multi-video alignment, which respectively correspond to the fields of view from two separate acquisition sessions. It can be clearly observed from the figure that there are deviations in the acquisition altitude and position between the two sessions.}
    \label{fig:align}
\end{figure}

In contrast to single-video stabilization, multi-video alignment is primarily aimed at unifying the field-of-view deviations caused by multiple acquisition sessions. Limited by the drone's battery life, each shooting session lasts approximately 30 minutes. For each session, it is necessary to manually re-align the reference position and altitude before reshooting. However, manual alignment makes it difficult to ensure the complete consistency of the field of view and flight altitude. Meanwhile, for subsequent map construction, unifying the coordinate systems of all videos enables the creation of a single map for each site. Therefore, multi-video alignment not only addresses the issue of alignment accuracy but also significantly improves the efficiency of map construction. For multi-video alignment, the method adopted in this study is to annotate multiple reference points for each video and calculate the transformation matrix using these reference points. The alignment effect is illustrated in the Figure \ref{fig:align}.

\subsection{Trajectory acquisition}
Trajectory acquisition is mainly realized through the sequential execution of object detection followed by multi-object tracking. The object detection results of each frame serve as the basis for multi-object tracking, so as to obtain complete and continuous target trajectories.
\subsubsection{Object detection}
Object detection and tracking is the process of acquiring the trajectories of all targets. In this study, YOLO11 \cite{yolo11_ultralytics} was adopted to implement oriented bounding box (OBB) detection for all targets (excluding pedestrians). A total of 3,000 images were annotated for training the YOLO11x-OBB model, which enables the detection of seven categories of targets, namely pedestrians, bicycles, motorcycles, tricycles, cars, trucks, and buses. The detection results were then used as the final detection bounding boxes after rectifying perspective transformation errors. The rectification method adopted in this study is the L-shape approach. It is based on the observation that perspective distortion is minimal near the center of the field of view in aerial images. The bounding boxes of the same target in other regions are rectified accordingly, thereby addressing the issue of bounding box overlapping under conditions of dense traffic flow. 

\subsubsection{Multi-object tracking}
For multi-object tracking, the ByteTrack \cite{zhang2022bytetrack} algorithm was employed in this study. ByteTrack fully leverages the results of object detection: it retains low-confidence detection results for a certain period as a backup, thereby preventing incomplete trajectories caused by low detection confidence at specific moments. To ensure the trajectory integrity of these small targets, additional processing was performed on small targets such as pedestrians, where the IoU matching threshold was adjusted. Figure \ref{fig:trackresults} presents the trajectory acquisition results.

\subsection{Data postprocessing and validation}
For data post-processing, we applied motion constraints to refine the trajectories while extracting multi-dimensional features to enhance the dataset representation. In this study, Rauch-Tung-Striebel (RTS) \cite{simon2006optimal} method was implemented for data post-processing. To enhance data quality and usability, and to minimize the learning and verification overhead for downstream researchers \cite{zhu2024generic}, we have developed and open-sourced a dedicated data rectification and visualization platform. Finally, we verified the data accuracy using a test vehicle equipped with RT inertial navigation equipment. 

\begin{figure*}[htbp]  
    \centering  
    \begin{subfigure}[b]{0.3\textwidth}
        \centering
        \includegraphics[width=\textwidth]{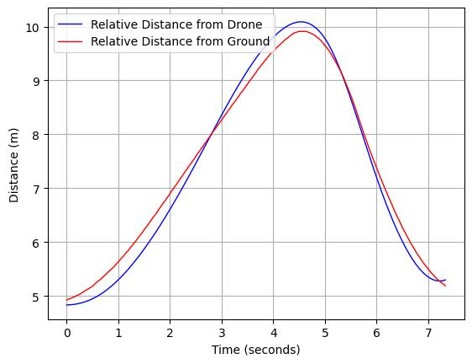} 
        \caption{}
        \label{fig:suba2}  
    \end{subfigure}
    \hfill 
    \begin{subfigure}[b]{0.3\textwidth}
        \centering
        \includegraphics[width=\textwidth]{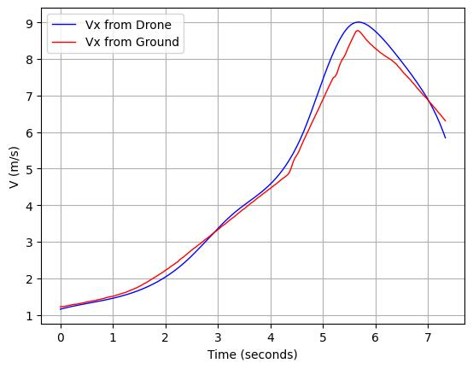} 
        \caption{}
        \label{fig:subb2}
        
    \end{subfigure}
    \hfill 
    \begin{subfigure}[b]{0.3\textwidth}
        \centering
        \includegraphics[width=\textwidth]{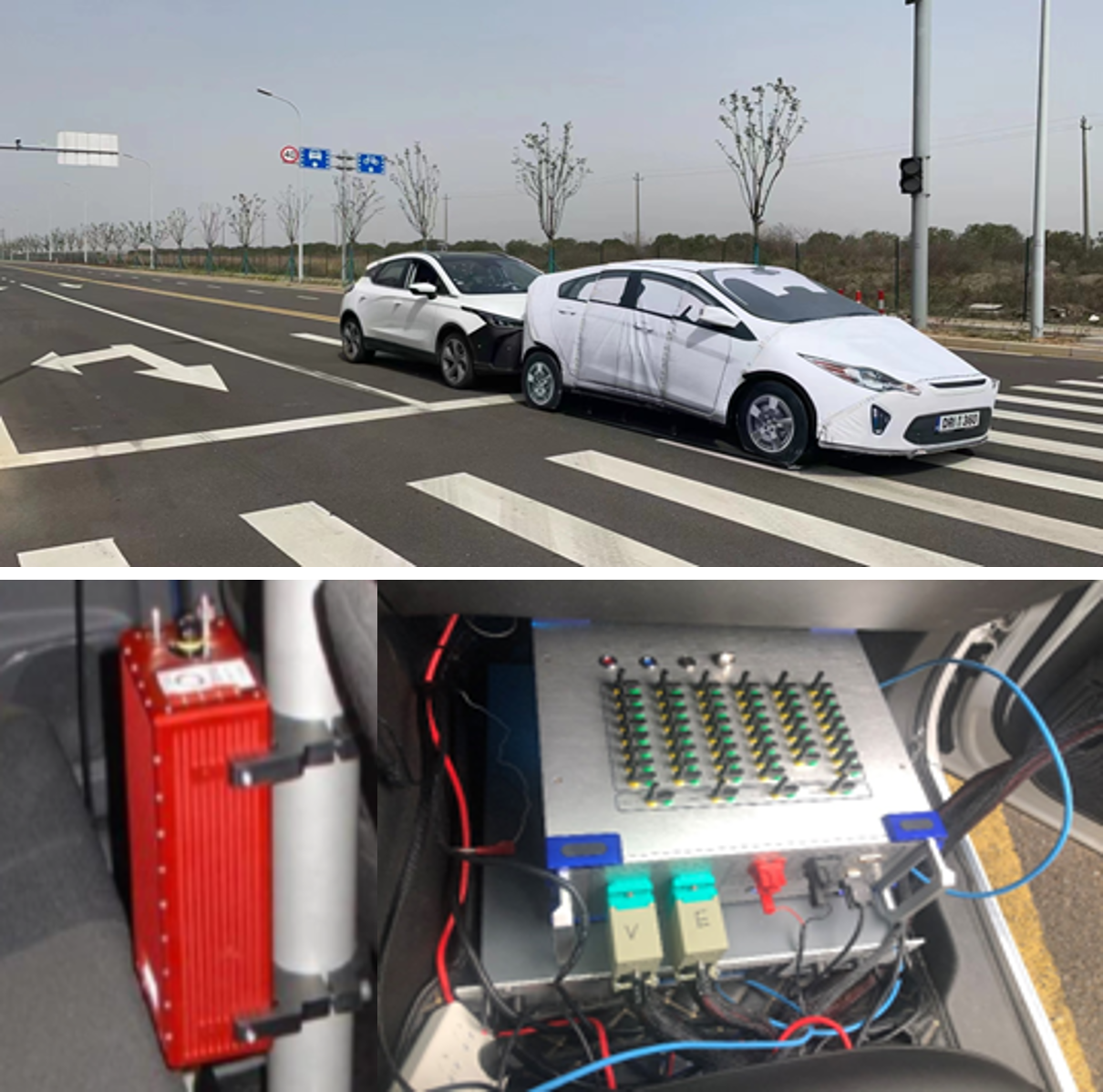} 
        \caption{}
        \label{fig:subc1}
    \end{subfigure}
    \caption{Data accuracy validation through vehicle-following experiments: 
    (a) Comparison of relative distance between the test vehicle and the soft target vehicle obtained from the drone-based extraction versus ground truth (RT inertial navigation). 
    (b) Comparison of the soft target vehicle's velocity ($V_x$) between the drone-extracted data and ground truth. 
    (c) Experimental setup showing the test vehicle chasing the soft target vehicle (top) and the RT-range high-precision inertial navigation equipment installed for ground truth acquisition (bottom).}
    \label{fig:valid}  

\end{figure*}

As shown in Figure \ref{fig:valid}, two vehicles were deployed in the test experiment: a test vehicle and a soft target vehicle. RT inertial navigation equipment was installed for calculating the distance and relative velocity between the two vehicles. In the experimental setup, the soft target vehicle was positioned ahead of the test vehicle; the soft target vehicle moved first, followed by the test vehicle performing a chasing maneuver. The relative distance between the two vehicles and the speed of the soft target vehicle were selected as the validation data types.

\subsection{Behavior definition and extraction}

\begin{figure}
    \centering
    \includegraphics[width=\linewidth]{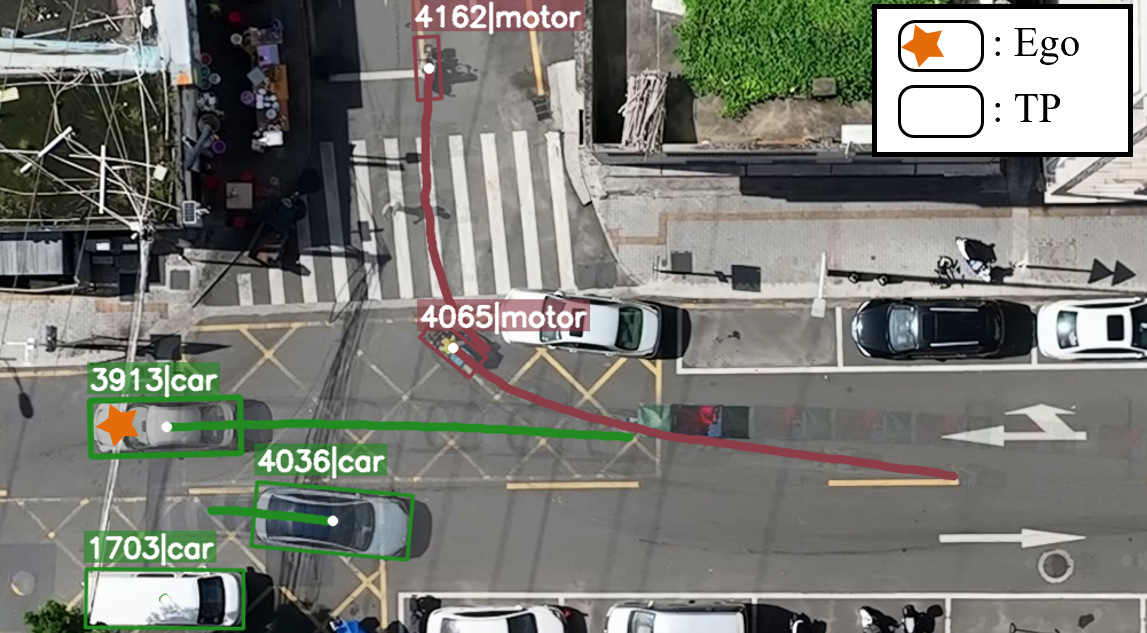}
    \caption{Multi-agent interaction scenario. The ego vehicle (ID 3913) interacts with highlighted critical targets. These agents exhibit a high tendency to approach the ego vehicle and must be prioritized for safe navigation through the intersection.}
    \label{fig:scenario_example}
\end{figure}

In mixed traffic environments, motor vehicles should navigate interactions with numerous surrounding traffic participants (TPs). To facilitate structured interaction analysis, we construct behavior samples in an ego-centered manner. Each behavior sample comprises the trajectory of the ego vehicle along with the trajectories of other TPs that exhibit potential interactions during the corresponding timeframe. Figure \ref{fig:scenario_example} illustrates a sample scenario of multi-agent interaction. The ego vehicle is designated as ID 3913, while the other highlighted objects represent traffic participants exhibiting a tendency to approach the ego vehicle. These extracted agents can be defined as critical targets that the ego vehicle must prioritize for safety and navigation as it maneuvers through the intersection. 

\begin{figure}[htbp]  
    \centering  
    \begin{subfigure}[b]{\linewidth}
        \centering
        \includegraphics[width=\linewidth]{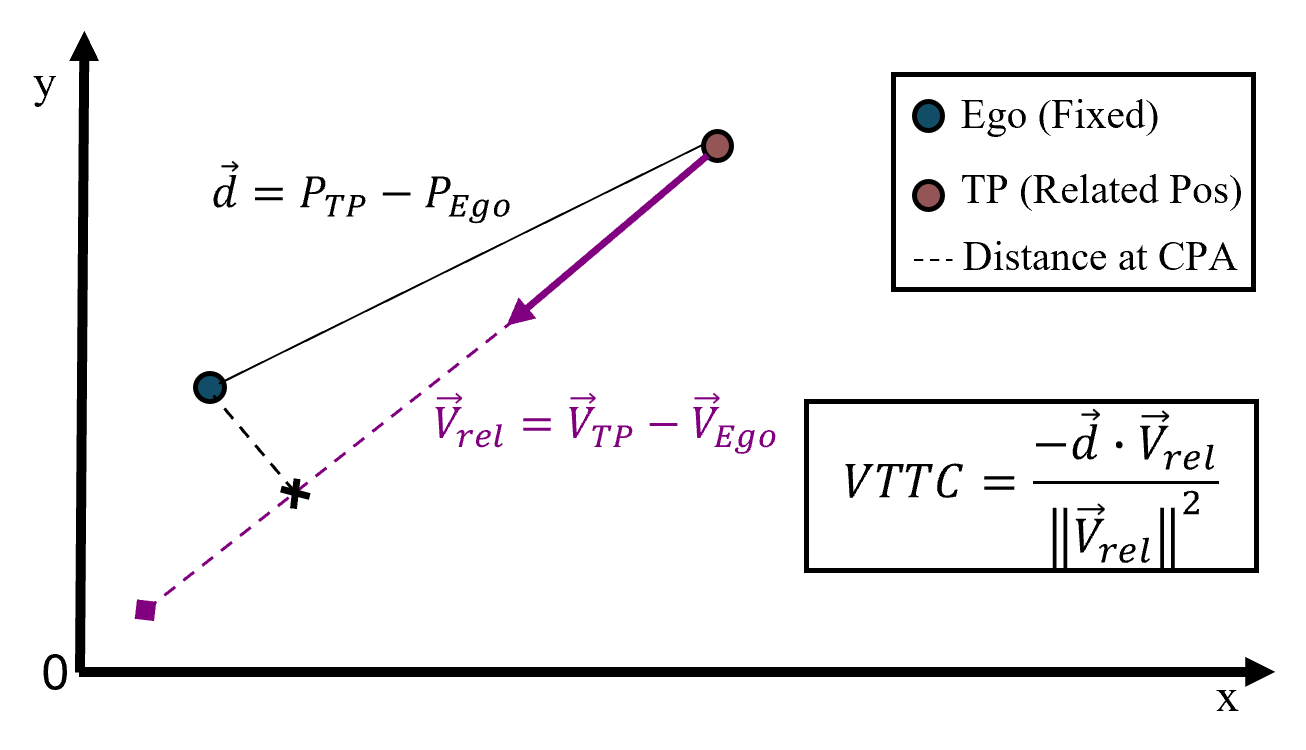} 
        \caption{}
        \label{fig:vttc2}  
    \end{subfigure}
    \hfill 
    \begin{subfigure}[b]{\linewidth}
        \centering
        \includegraphics[width=\linewidth]{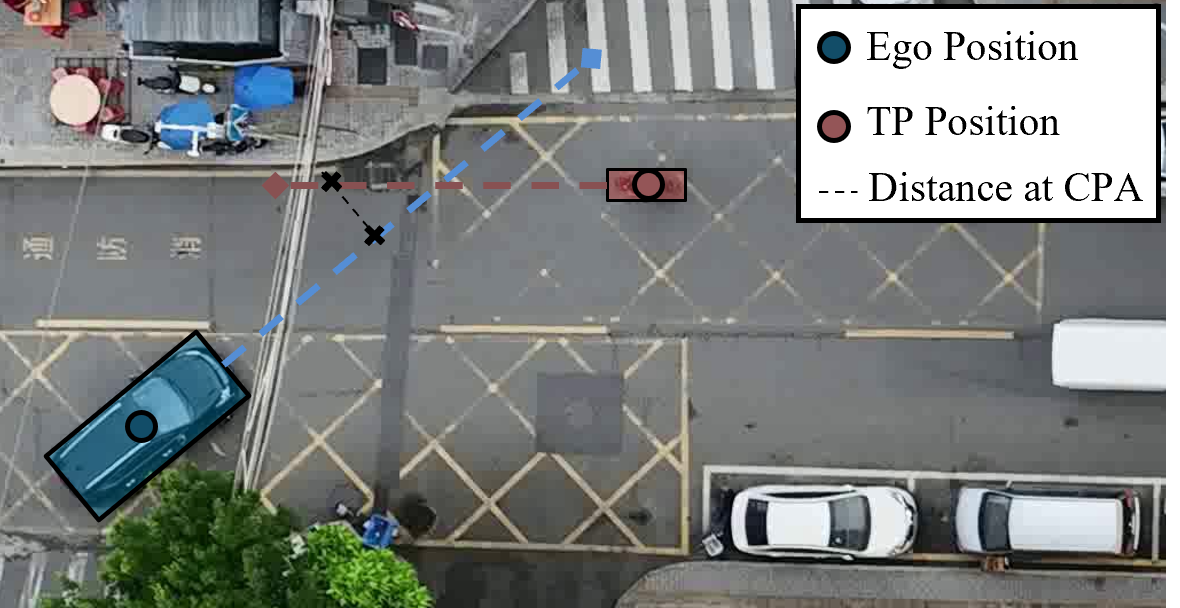} 
        \caption{}
        \label{fig:vttc1}
    \end{subfigure}
    \caption{The calculation formula and a real-world example of VTTC are shown in Fig. (a) and Fig. (b), respectively. This metric quantifies the temporal proximity of the TP to the Ego at a specific instance.}
    \label{fig:VTTC} 
\end{figure}

To quantify interaction relevance, we introduce a novel Surrogate Safety Measure (SSM) \cite{wang2021review} termed Vector Time to Collision (VTTC). Developed as an enhancement of the Time to Collision (TTC) \cite{minderhoud2001extended}, VTTC utilizes velocity vectors and the Closest Point of Approach (CPA) to calculate the degree of proximity between the ego and other TPs, thereby quantifying interaction relevance. Furthermore, by analyzing the statistical distribution of VTTC, we set a threshold for interaction relevance to filter and extract significant interaction behaviors. Figure \ref{fig:VTTC} presents the calculation formula and a corresponding real-world example of VTTC.

To extract representative samples of multi-agent interaction scenarios, we implemented a comprehensive traversal strategy. Each motorized vehicle within a scene is iteratively designated as the ego vehicle, and the VTTC is calculated between the ego and its TPs. Interacting TPs are then identified by applying a predefined VTTC threshold.

Mathematically, VTTC is represented as a positive value, where proximity to zero signifies a higher intensity of interaction. To establish a data-driven filtering threshold for quantifying these interactions, we performed a statistical analysis on the mean VTTC of each initially screened scenario. A box plot was subsequently employed to visualize the distribution of these mean values, facilitating the determination of an optimal VTTC threshold that balances scenario complexity with interaction significance.

\begin{figure}
    \centering
    \includegraphics[width=\linewidth]{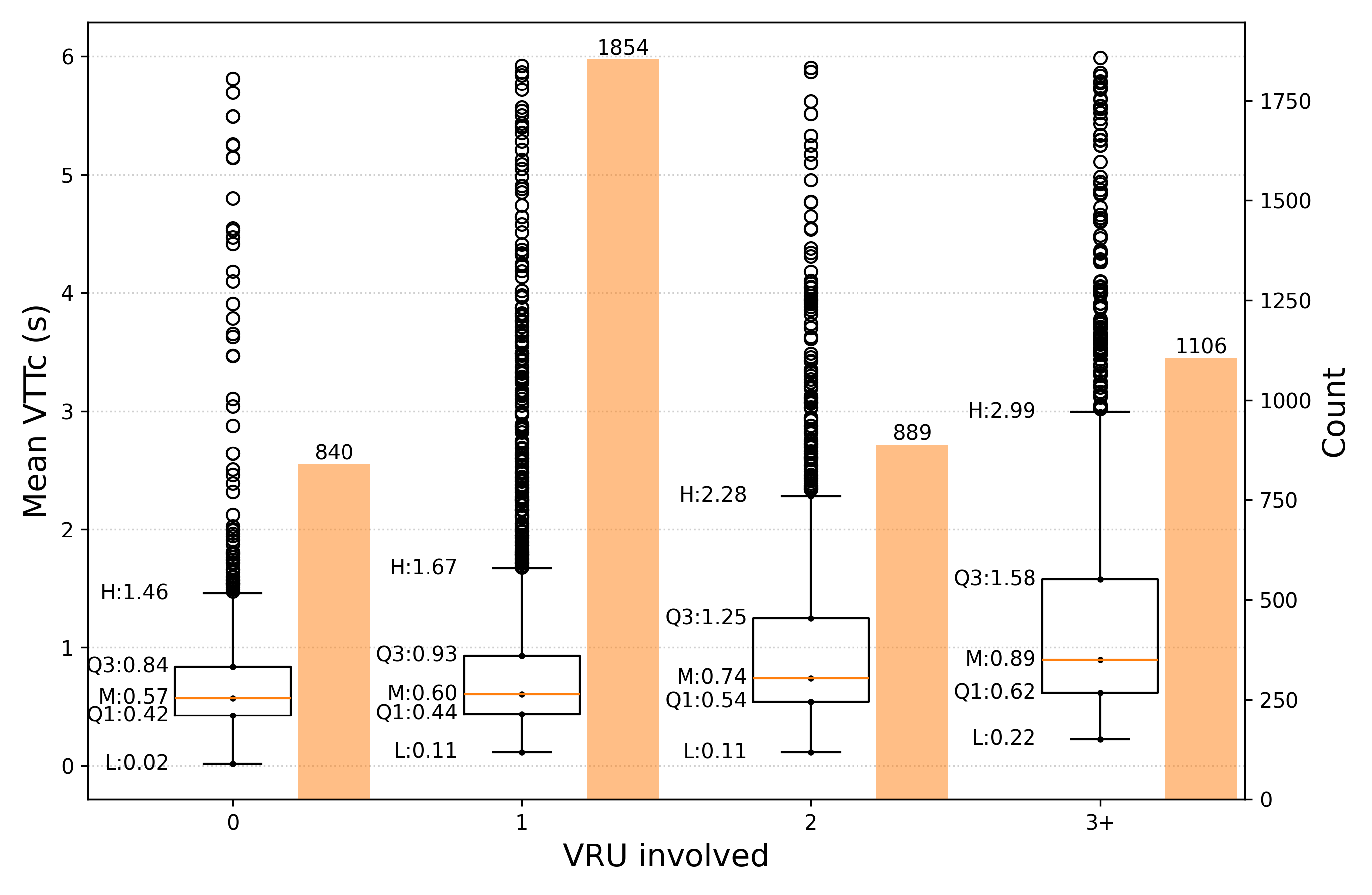}
    \caption{Observing the positive correlation between VRU count, complexity, and mean VTTC.}
    \label{fig:VTTC_box}
\end{figure}

Figure \ref{fig:VTTC_box} shows that the mean VTTC increases with VRU density, suggesting that vehicles may adopt more conservative driving behaviors in higher-density scenarios. To balance risk identification with data retention, the upper quartile (Q3) value of 1.53 s (derived from the most complex scenarios) was adopted as the cut-off. This strategy allows for the maximization of complex scenario retention while eliminating non-interactive noise associated with large VTTC values.

\subsection{Data formats}
In this paper, the VRUD dataset is structured into three components: (1) road user trajectories, (2) high-definition map data, and (3) interaction behavior data.
\begin{itemize}
    \item \textbf{Trajectory data:} The trajectory data encompasses both static and dynamic attributes of traffic participants. Static attributes include target dimensions, position, category, and heading angle, while dynamic attributes comprise velocity, acceleration, and yaw rate. All data is unified within the image coordinate system and stored in CSV format. Notably, the data sampling frequency is synchronized with the original aerial video frame rate at 30 Hz.
    \item \textbf{Map data:} The map data package primarily comprises high-definition maps in the standard OpenDRIVE format, accompanied by high-resolution aerial base maps and map registration documentation. To ensure synchronization between the image coordinate system and the global coordinate system, we performed precise calibration and provided the necessary conversion coefficients and matching parameters. This facilitates the accurate alignment of trajectory data with the map files. Furthermore, to guarantee data security, all actual geographic information has been anonymized (de-identified), preserving only the relative spatial relationships required for research.
    \item \textbf{Interaction behavior data:} The interaction behavior file serves as a comprehensive scenario index, facilitating the efficient extraction of scenario samples (as defined in this study) from the raw trajectory data. Key indexing information includes the Ego-vehicle ID, the start and end frames (temporal window), and the identifiers of other relevant traffic participants.
\end{itemize}

\begin{figure}
    \centering
    \includegraphics[width=\linewidth]{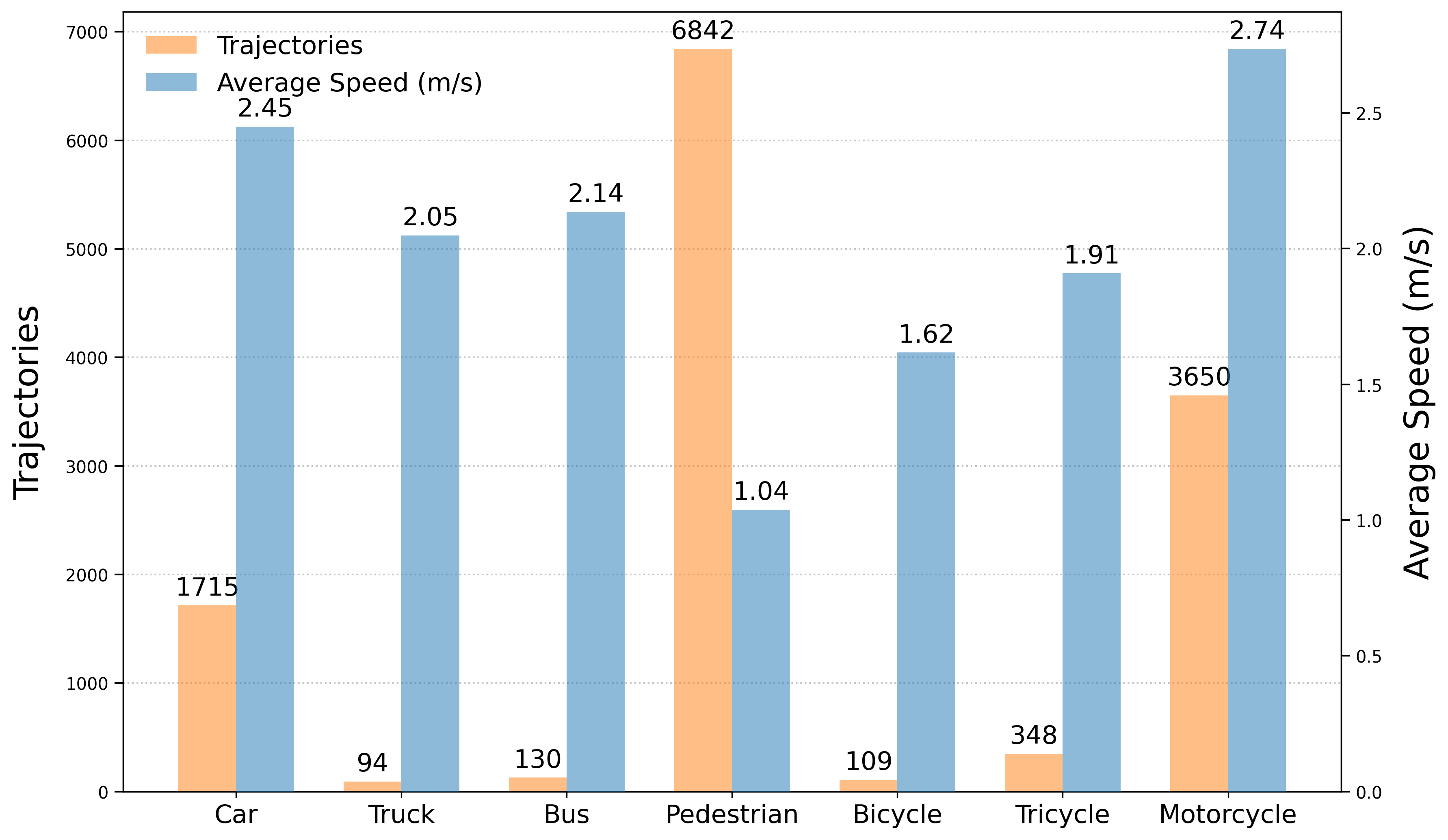}
    \caption{Categorical distribution and average velocity statistics. The results highlight the predominance of pedestrians and motorcycles in count, while the velocity distribution underscores the high traffic efficiency of motorcycles in unstructured environments.}
    \label{fig:class_bar}
\end{figure}

\section{Dataset Statistics and Behavioral Characterization}
\label{sec:statistics}
\begin{figure*}[t] 
    \centering
    
    \begin{subfigure}[b]{0.48\linewidth} 
        \centering
        \includegraphics[width=0.9\linewidth]{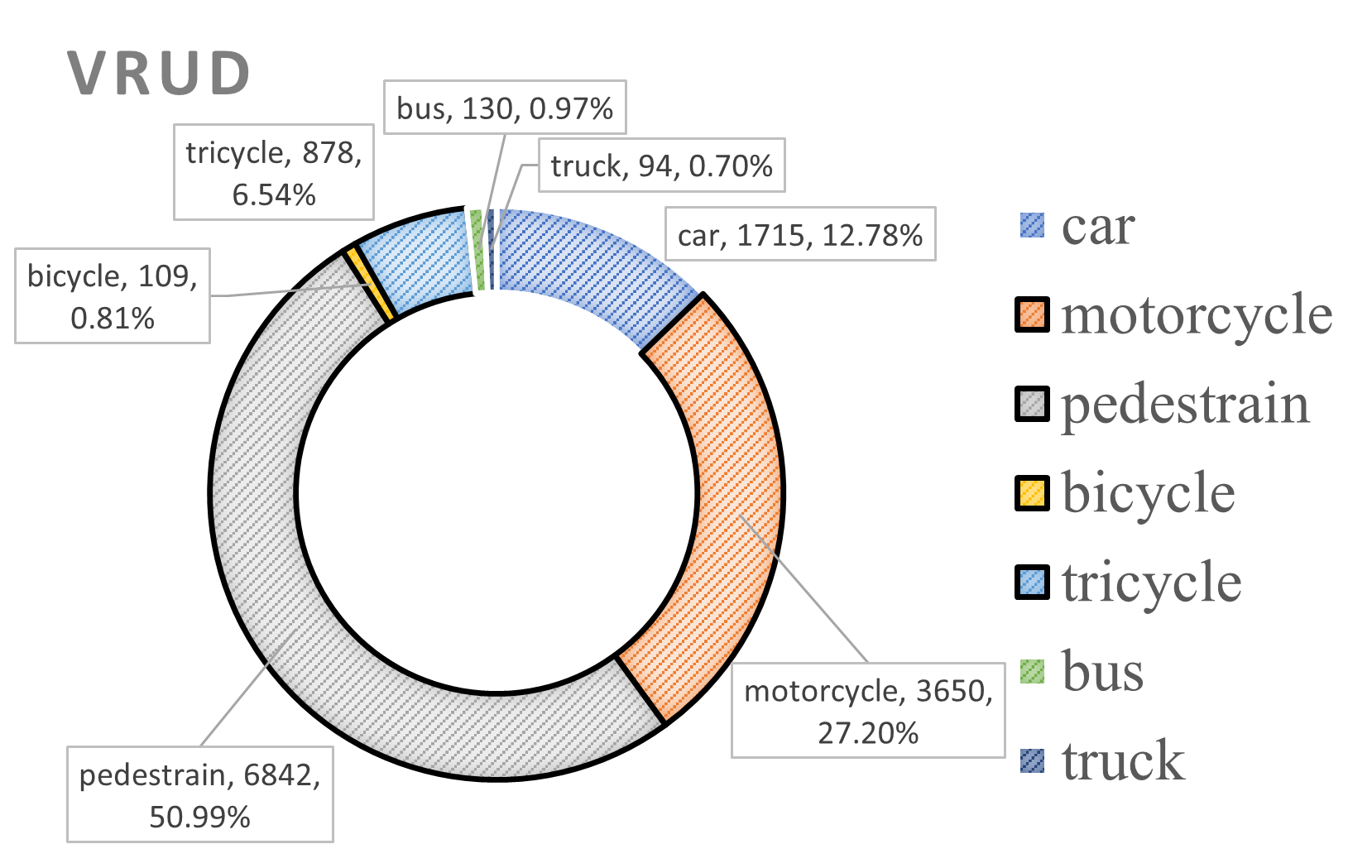}
        \caption{}
        \label{fig:suba3}
    \end{subfigure}
    \hfill 
    \begin{subfigure}[b]{0.48\linewidth}
        \centering
        \includegraphics[width=0.9\linewidth]{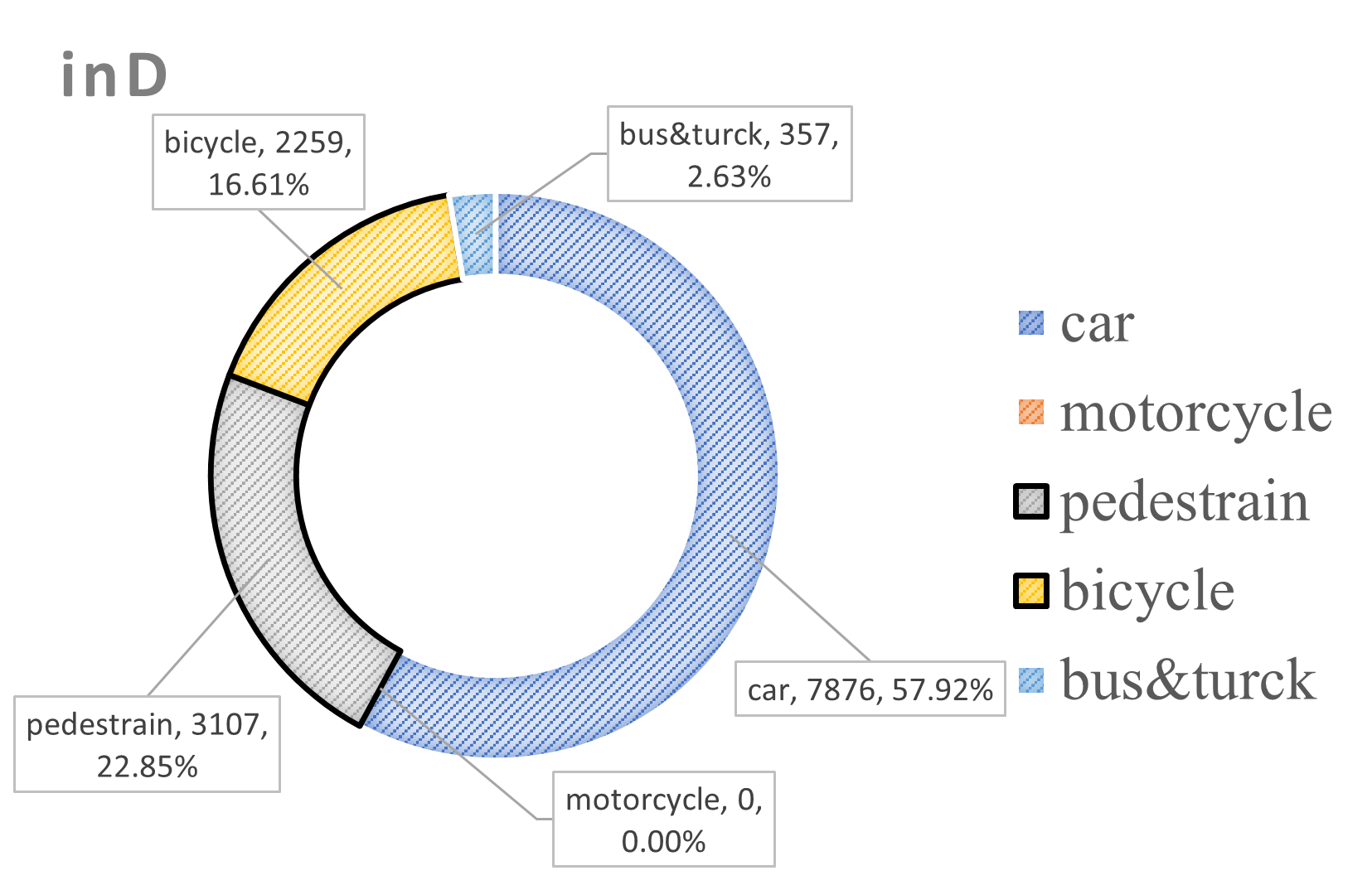}
        \caption{}
        \label{fig:subb3}
    \end{subfigure}
    
    \vspace{1em} 
    
    \begin{subfigure}[b]{0.48\linewidth}
        \centering
        \includegraphics[width=0.9\linewidth]{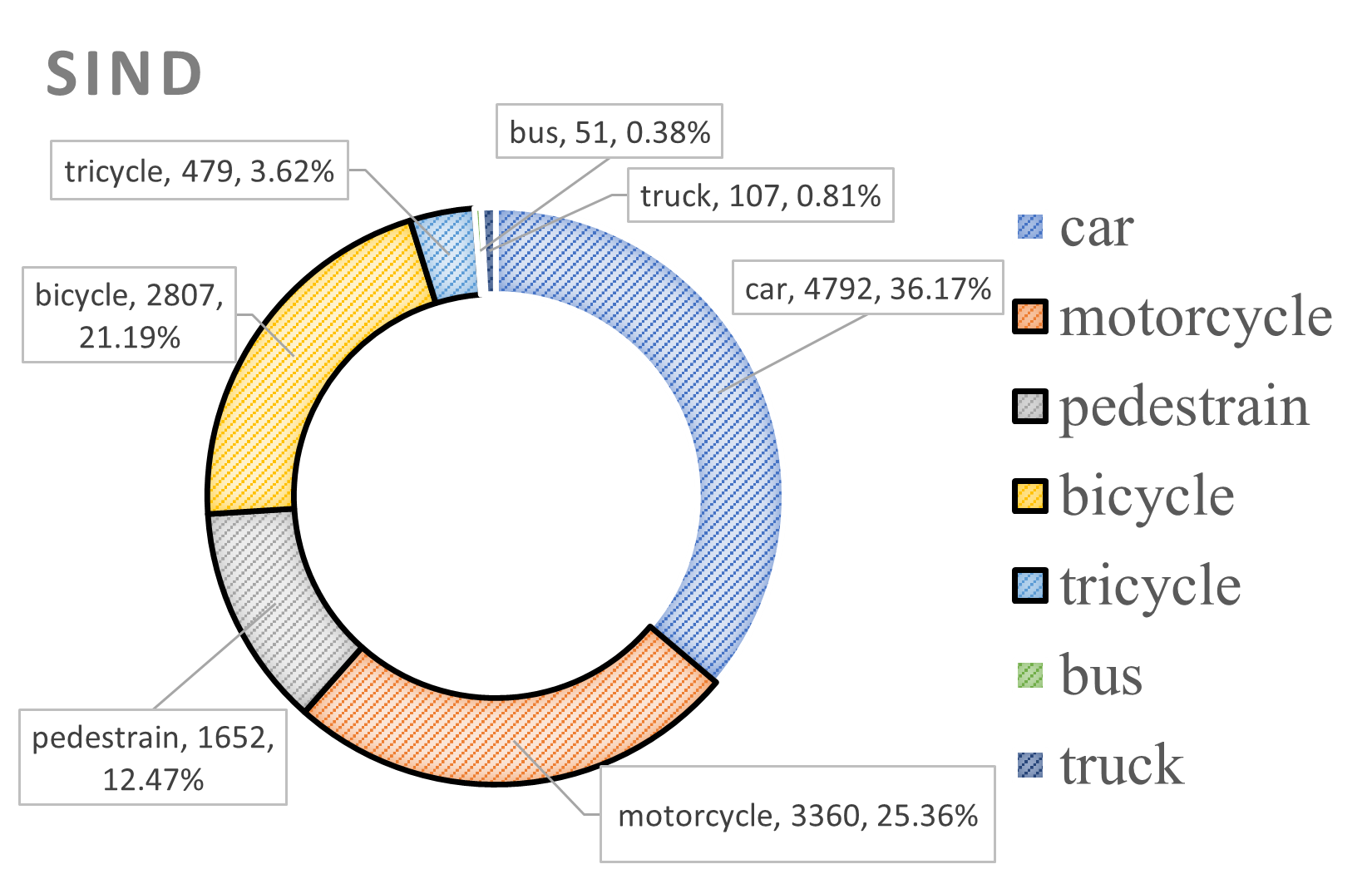}
        \caption{}
        \label{fig:subc3}
    \end{subfigure}
    \hfill
    \begin{subfigure}[b]{0.48\linewidth}
        \centering
        \includegraphics[width=0.9\linewidth]{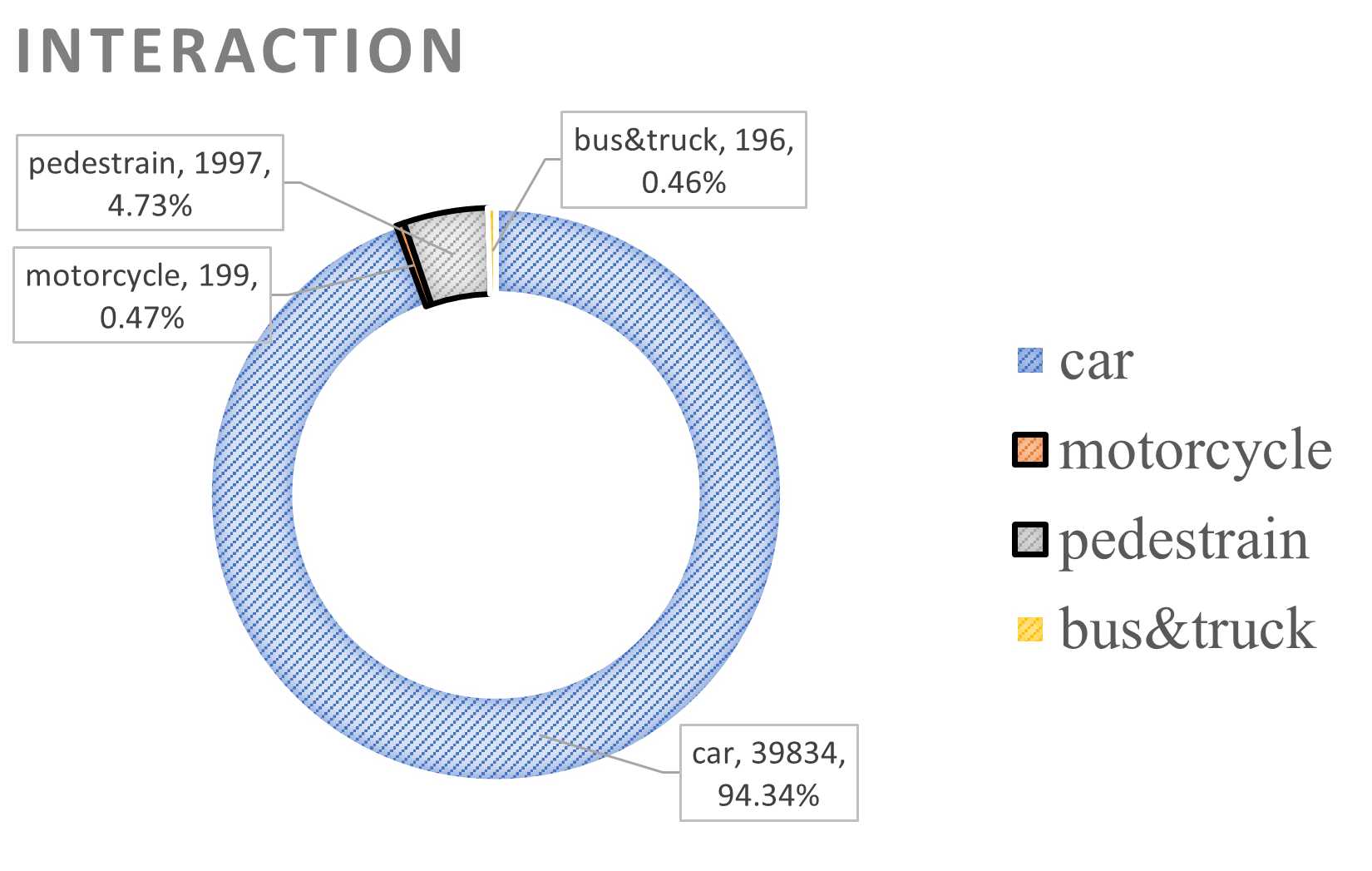}
        \caption{}
        \label{fig:subd3}
    \end{subfigure}
    
    \caption{The figure compares the data distributions of VRUD with those of the inD, SIND, and INTERACTION datasets; it is clearly observable that VRUs account for nearly half of the VRUD dataset, significantly exceeding the proportions found in other datasets.}
    \label{fig:main_figure}
\end{figure*}
This section presents an overview of the VRUD dataset and provides a statistical analysis of the extracted behaviors. It aims to enable interested researchers to gain immediate insights into the specialized processing applied to VRUs within this open-source dataset.

\subsection{Dataset Statistics}

The VRUD dataset was collected from two typical urban villages in Shenzhen, China, comprising 4 hours of effective data and 12,888 trajectories. It encompasses seven categories of road users: cars, buses, trucks, bicycles, motorcycles, tricycles, and pedestrians. As shown in Table \ref{tab: comparison}, we compare our dataset with three widely adopted open-source datasets that involve VRUs: SIND, inD, and INTERACTION. VRUD is characterized by a dominant proportion of VRUs, accounting for nearly 85\% of all trajectories (51\% pedestrians, 27\% motorcycles, a collective 7\% for tricycles and bicycles). 

To analyze the spatial characteristics, Figure \ref{fig:trajectories} presents that all trajectories were projected onto a unified base map. The visualization reveals distinct movement patterns: pedestrian trajectories are highly dispersed; motor vehicle paths are significantly constrained by roadside parking; whereas motorcycle trajectories exhibit a nearly random distribution, covering the entire navigable road surface. 

Furthermore, in Figure \ref{fig:class_bar}, velocity statistics indicate that motorcycles possess the highest average speed in these mixed traffic scenarios, followed by cars, highlighting the superior mobility efficiency of motorcycles in mixed traffic.

\begin{table*}[!ht]
    \centering
    \caption{Comparison of characteristics among open-source datasets}
    \label{tab: comparison}
    \begin{tabular}{|l|l|l|l|p{6.5cm}|l|l|l|}
    \hline
        Dataset name & Length & Trajectories & Density & Road user types & HD Map & Sample Freq  \\ \hline
        INTERACTION & 16.5 h & 40054 & 2427.5 TPs/h & Pedestrian/bicycle, and car & lanelet2 & 10 Hz   \\ \hline
        InD & 10.0 h & 13599 & 1359.9 TPs/h & Pedestrian, bicycle, car, and bus & lanelet2 & 25 Hz   \\ \hline
        SIND & 7.0 h & 13248 & 1892.6 TPs/h & Car, bus, truck, bicycle, motorcycle, tricycle, pedestrian & lanelet2 & 10 Hz  \\ \hline
        VRUD (ours) & 4.0 h & 12888 & \textbf{3222 TPs/h} & Car, bus, truck, bicycle, motorcycle, tricycle, pedestrian & \textbf{OpenDRIVE} & \textbf{30 Hz}   \\ \hline
    \end{tabular}
    \vspace{1em} 
    \footnotesize \textit{Note}: TPs denotes traffic participants. 
\end{table*}

\begin{figure}
    \centering
    \includegraphics[width=\linewidth]{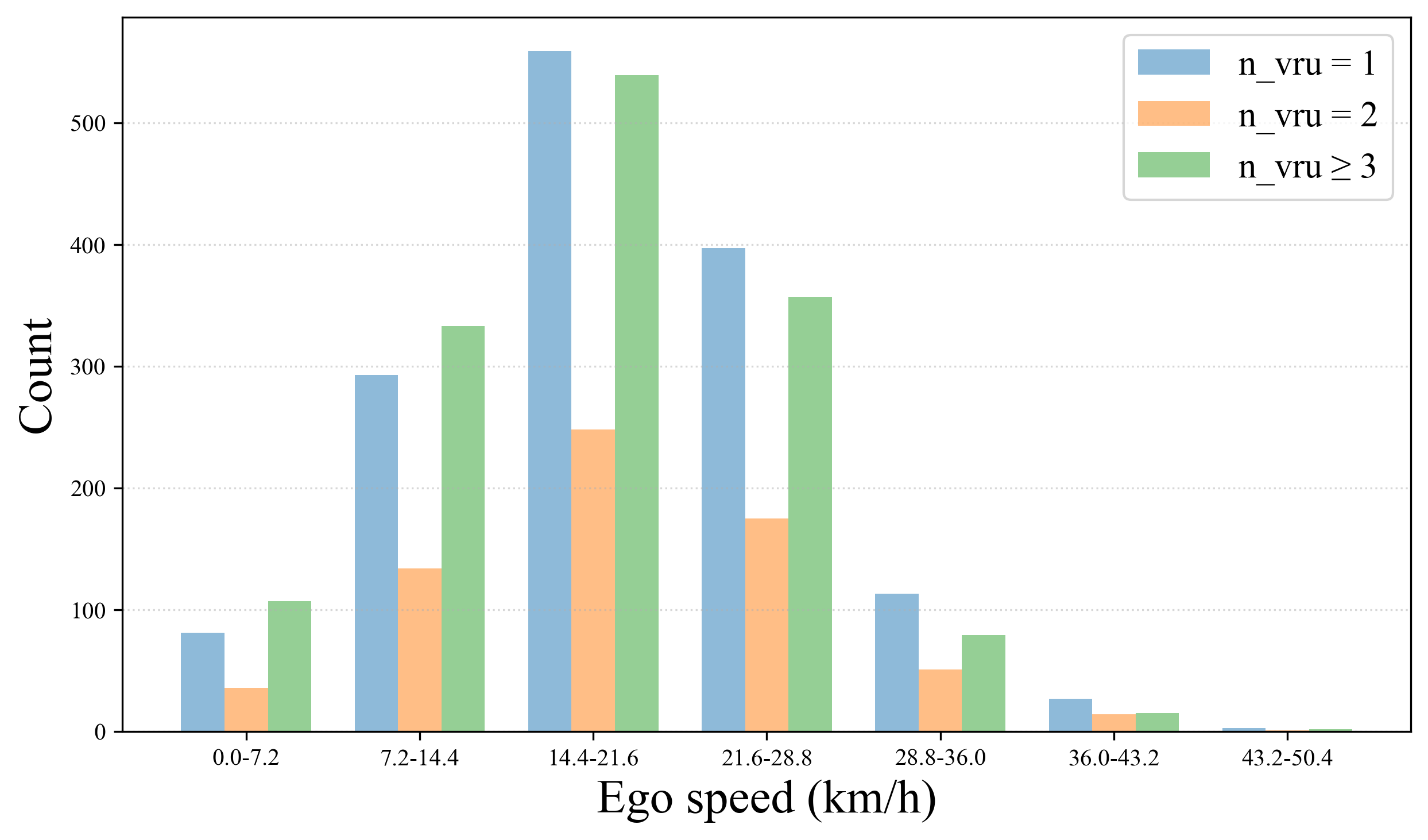}
    \caption{Empirical ego-speed distribution relative to VRU counts. The quasi-normal distribution of VRUD samples across ego-velocity and VRU density highlights the characteristic low-speed maneuvering of vehicles in urban mixed traffic scenarios.}
    \label{fig:ego_V_vru_count}
\end{figure}
To further investigate the interaction characteristics between cars and distinct categories of VRUs, we conducted a statistical analysis of scenario samples stratified by VRU type. Specifically, we examined the correlations among VRU velocity, vehicle velocity, and VTTC.
\begin{figure}
    \centering
    \includegraphics[width=\linewidth]{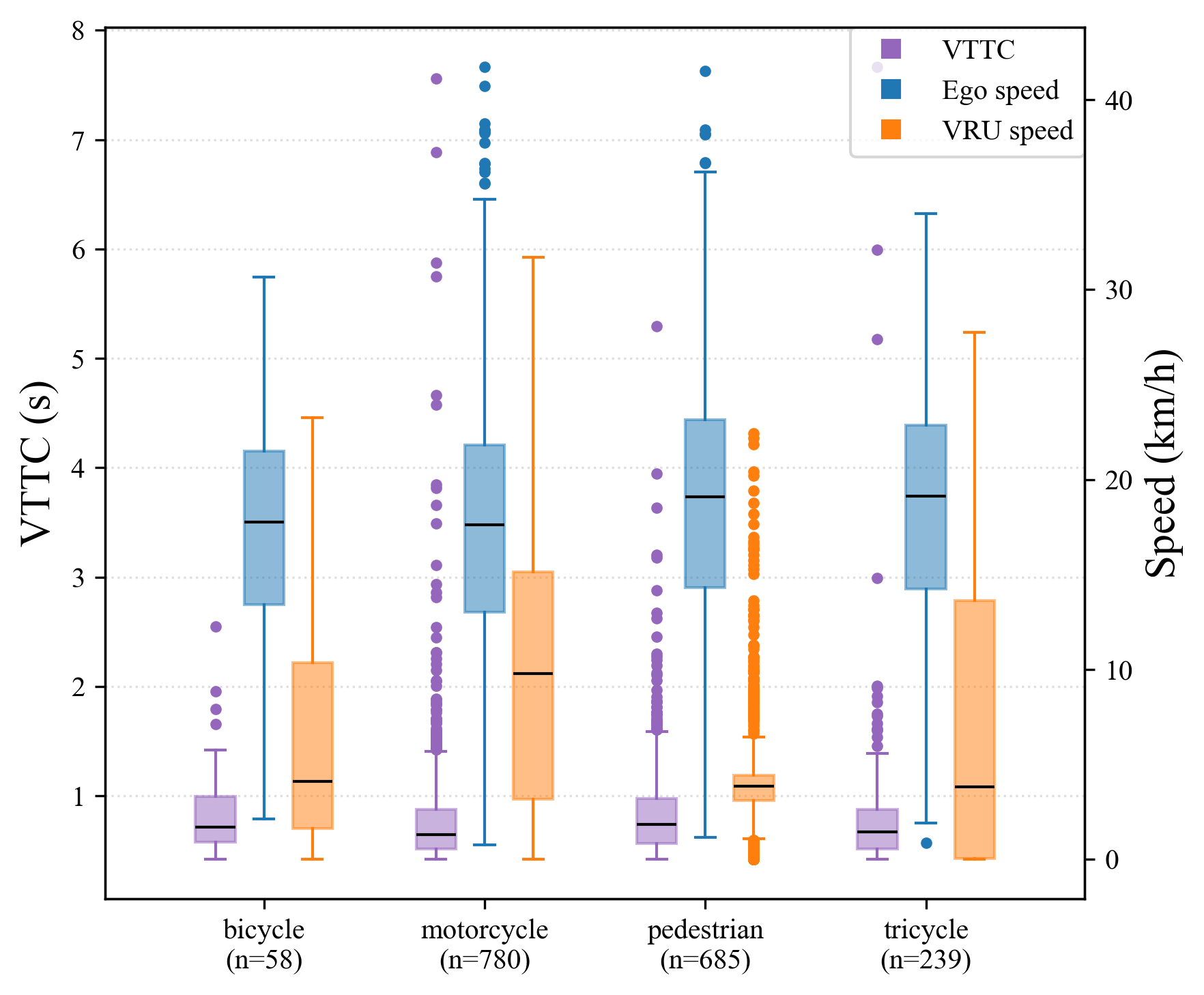}
    \caption{Characterization of interaction intensity via velocity-VTTC coupling. The ego-vehicle maintains a tactical velocity corridor (17.5–20.0 km/h) while VTTC consistently converges around 0.7s. The 0.7s mark establishes a robust quantitative filter for extracting high-value, interaction-critical samples from heterogeneous traffic data.}
    \label{fig:VTTC_0.7}
\end{figure}

\subsection{Behavioral Characterization}
Following the application of the VTTC filtering threshold, a final set of 4,002 valid scenarios was retained from an initial pool of approximately 5,000 samples. Within this dataset, scenarios featuring a car as the Ego-vehicle are predominant, accounting for 3,600 cases. Consequently, our subsequent analysis focuses exclusively on these car-based scenarios. 

To investigate the interaction dynamics in mixed traffic environments, we performed a joint statistical analysis categorizing the samples based on two dimensions: the Ego-vehicle's velocity and the number of surrounding VRUs. Figure \ref{fig:ego_V_vru_count} exhibits a quasi-normal distribution across the velocity space, with the primary operational range concentrated between $14.4$ and $28.8 \text{ km/h}$. This interval represents the equilibrium between operational efficiency and safety constraints within mixed traffic environments. A significant inverse correlation is observed between VRU density and ego-vehicle speed. In high-density scenarios ($n\_vru \ge 3$), the distribution shifts toward lower velocity brackets ($0.0 - 14.4 \text{ km/h}$), indicating a proactive risk-mitigation strategy where the ego-vehicle increases safety margins in response to increased environmental complexity.

Figure \ref{fig:VTTC_0.7} illustrates that the ego-vehicle maintains a tactical velocity corridor (17.5–20.0 km/h) to manage latent conflicts. Crucially, the VTTC distribution, consistently clustered around 0.7s, serves as a proxy for interaction relevance rather than a direct metric of collision risk. In this context, the magnitude of VTTC indicates the intensity of spatio-temporal coupling: a lower value signifies a highly relevant interaction requiring active negotiation, not necessarily an imminent safety violation. This convergence underscores a universal protocol where heterogeneous kinematics are homogenized into a stable temporal threshold for interaction activation at the CPA. 

Ultimately, the identification of the 0.7s invariance serves as a critical takeaway from this study. By defining this value as a robust indicator of interaction relevance, we provide a quantitative filter for subsequent research. This work lays the groundwork for future investigators to bypass the noise of non-interacting agents and directly target high-value samples. We envision this 0.7s threshold becoming a plug-and-play parameter for extracting 'interaction-critical' subsets, significantly streamlining the training and validation processes for next-generation autonomous systems.

\section{Conclusion}
\label{sec:conclusion}
This paper presents a new dataset named VRUD. As autonomous driving technology advances toward Level 4 and beyond, modeling interactions with VRUs in mixed urban traffic has emerged as a critical bottleneck. The open-source release of VRUD is specifically aimed at addressing the severe scarcity of high-quality, real-world trajectory data within this specialized domain for both academia and industry.

The core value of this dataset is reflected in the following three dimensions:
\begin{itemize}
    \item \textbf{Supplementing Rare Edge Cases:} Unlike existing datasets focused on highways or regulated intersections, VRUD captures non-structured mixed traffic environments such as urban villages. These scenarios are characterized by extreme VRU density, stochastic interaction behaviors, and complex occlusions, representing the most challenging ``long-tail'' edge cases for current Autonomous Driving Systems.
    \item \textbf{High-Dimensional Open-Sourcing of Full-Element Trajectories:} Beyond high-precision trajectory data, we provide a comprehensive suite of supporting resources, including HD maps in OpenDRIVE format and trajectory calibration parameters. The data encompasses seven categories of traffic participants. These range from pedestrians and bicycles to regionally characteristic motorcycles and tricycles, providing a foundational pillar for multi-agent trajectory prediction and behavioral intention recognition.
    \item \textbf{Standardized Extraction of Large-Scale Scenario Libraries:} Moving beyond a simple aggregation of raw traffic flow, VRUD provides a refined library of 4,002 multi-agent interaction scenario samples. This transformation from bulk trajectory data to a structured scenario database significantly lowers the barrier for data preprocessing, enabling direct support for the training and simulation-based validation of end-to-end decision-making models.
\end{itemize}

The comprehensive open-sourcing of VRUD aims to catalyze research iterations in the safety and socio-collaborative performance of autonomous vehicles by sharing authentic, complex interaction data. We cordially invite the global research community to explore the potential of this dataset and join us in overcoming the challenges of deploying autonomous driving in complex urban environments.

{
\fontsize{8}{9}\selectfont 
\bibliographystyle{IEEEtran}
\bibliography{ref}
}

\end{document}